\documentclass[a4paper,fleqn]{cas-dc}

\usepackage[numbers]{natbib}

\def\tsc#1{\csdef{#1}{\textsc{\lowercase{#1}}\xspace}}
\tsc{WGM}
\tsc{QE}
\tsc{EP}
\tsc{PMS}
\tsc{BEC}
\tsc{DE}

\usepackage{algorithm}
\usepackage{algorithmic}
\usepackage{chngcntr}
\begin{document}
\let\WriteBookmarks\relax
\def\floatpagepagefraction{1}
\def\textpagefraction{.001}
\shorttitle{Analysing Moral Bias in Finetuned LLMs through Mechanistic Interpretability}
\shortauthors{B. Raimondi et~al.}

\title [mode = title]{Analysing Moral Bias in Finetuned LLMs through Mechanistic Interpretability}                      

\author[1]{Bianca Raimondi}[orcid=0009-0002-1562-7722]
\cormark[1]
\ead{bianca.raimondi3@unibo.it}
\credit{Conceptualization, Data curation, Formal analysis, Methodology, Software, Writing – original draft
}

\affiliation[1]{organization={Department of Computer Science and Engineering, University of Bologna},
                city={Bologna},
                country={Italy}}

\author[2]{Daniela Dalbagno}[orcid=0000-0002-4027-8942]
\ead{daniela.dalbagno2@unibo.it}
\credit{Conceptualization, Data curation, Writing – original draft
}

\author[1]{Maurizio Gabbrielli}[orcid=0000-0003-0609-8662]
\ead{maurizio.gabbrielli@unibo.it}
\credit{Supervision}

\affiliation[2]{organization={Center for Studies and Research in Cognitive Neuroscience, Department of Psychology, University of Bologna},
                city={Cesena},
                country={Italy}}

\cortext[cor1]{Corresponding author}

\begin{abstract}
Large language models (LLMs) have been shown to internalize human-like biases during finetuning, yet the mechanisms by which these biases manifest remain unclear. In this work, we investigated whether the well-known Knobe effect, a moral bias in intentionality judgements, emerges in finetuned LLMs and whether it can be traced back to specific components of the model. We conducted a Layer-Patching analysis across 3 open-weights LLMs and demonstrated that the bias is not only learned during finetuning but also localized in a specific set of layers. Surprisingly, we found that patching activations from the corresponding pretrained model into just a critical layer is sufficient to eliminate the effect. Our findings indicate that social biases in LLMs can be interpreted, localized, and mitigated through targeted interventions, without the need for model retraining.
\end{abstract}


\maketitle

\section{Introduction}\label{sec:introduction}

An essential component of human moral judgement is the ability to attribute mental states, such as beliefs and intentions, to moral agents \cite{young_2007}. However, the intentionality attribution process is susceptible to cognitive biases \cite{schramowski2022large, o2025morality}. A well-documented example is the Knobe effect, which demonstrates that individuals are more likely to judge an action as intentional when it leads to a negative side effect than when it results in a positive one, even if the agent’s intention remains the same \cite{knobe2003intentional}.

As Large Language Models (LLMs) become increasingly integrated into decision-making and ethical reasoning tasks~\cite{dubey2025addressingmoraluncertaintyusing, dillion2025ai}, it is crucial to investigate whether they also exhibit the Knobe effect. Understanding this can shed light on how closely their moral reasoning aligns with human cognition and values.

In this study, we addressed three key research aims.
First (RQ1), we examined whether different LLMs, specifically Llama~\cite{grattafiori2024llama}, Mistral~\cite{jiang2023mistral7b}, and Gemma~\cite{team2024gemma}, manifest the Knobe effect in a way comparable to human behavior.

Consistent with prior findings~\cite{itzhak2024instructed, raimondi2025exploiting}, our work confirmed that these models systematically reproduce human biases, particularly after finetuning. Although designed to align models with human preferences, finetuning also causes the internalization of human biases, which can lead to unintended answers.

Second (RQ2), we investigated the internal mechanisms through which such bias is encoded, with a specific focus on the Transformer layers.
Our analysis revealed that the Knobe effect tends to be localized in mid-to-late layers, suggesting potential leverage points for bias mitigation.

Finally (RQ3), we explored whether the observed bias can be effectively reduced.
To this end, we developed a Layer-Patching algorithm that selectively overwrites internal activations of finetuned models using those from their pretrained version.
Our findings indicate that this method reduces the bias without altering model parameters or further training.

Overall, our work follows these research questions:
\begin{itemize}
\item \textbf{RQ1}: Do LLMs exhibit the Knobe effect? If so, to what extent is it a consequence of finetuning?
\item \textbf{RQ2}: Are specific Transformer layers responsible for encoding this bias?
\item \textbf{RQ3}: Can targeted interventions, such as Layer-Patching, be used to mitigate or eliminate this bias?
\end{itemize}

\section{Related Work}\label{sec:related_work}

Research on the Knobe effect \cite{knobe2003intentional} revealed that intentionality attribution is shaped by the valence of an action’s outcome, highlighting that such judgements are not purely objective but influenced by evaluative processes. Building on this, substantial psychological and neuroscientific research has linked impairments in intentionality attribution (due to brain lesions, neurodevelopmental conditions, or psychiatric disorders) to disruptions in social functioning and moral judgement \cite{baez_2014, Brune_2005, georges_2013, young_2010, sarfati_1997, Starita_2025, zucchelli_2019}.

While these studies focus on human cognition, recent work has begun to examine whether Artificial Intelligence (AI) systems, particularly LLMs, exhibit similar moral asymmetries. \citet{slobodenyuk2024moral} demonstrates that LLMs display moral asymmetries in ethically charged scenarios, suggesting possible parallels between model outputs and human moral reasoning. Recent empirical investigations have further mapped this landscape; for instance, \citet{ding2025pull} benchmarked moral biases across leading LLMs, while \citet{abdulhai2024moral} explored how diverse moral foundations are internally represented, providing a necessary theoretical precursor to our mechanistic study. In a broader context, \citet{bender2021dangers} and \citet{weidinger2022taxonomy} have raised concerns about how LLMs may absorb and perpetuate social biases from training data, including those related to morality. \citet{turpin2023language} further investigated how finetuning strategies can amplify or suppress such biases, pointing to the training as a key factor influencing model alignment. Crucially, \citet{kiehne2024analyzing} directly investigated how learning downstream tasks through finetuning alters moral bias, aligning with our focus on how alignment procedures can inadvertently amplify pre-existing cognitive asymmetries \cite{cheung2025large}.

However, direct analyses of whether LLMs replicate the Knobe effect remain scarce. This leaves open fundamental questions about the alignment between human and machine moral judgements, forming the basis for our RQ1.

Beyond behavioral evaluations, recent work has explored structural and mechanistic differences between human cognition and LLMs. \citet{collacciani2024quantifying} analyzed divergences in reasoning patterns, while \citet{bonard2024improving} investigated how affective signals can be integrated into model responses to improve alignment with human expectations. These contributions highlight that human-like moral reasoning in LLMs depends not only on data but also on how internal representations are shaped.

On a more philosophical level, \citet{chalmers2023could} questioned whether LLMs can possess intentionality or simulate it via statistical correlations. This debate underscores the ethical and epistemological importance of understanding biases in LLMs, not merely as artifacts of training data but as emergent properties of model architecture.

While much of the prior work focused on detecting biases at the behavioral level, few studies address where such biases are represented within model architectures or how they might be removed. Advances in mechanistic interpretability\cite{yu2026tracing}, such as Layer-Patching~\cite{meng2022locating,conmy2023automated,nanda2022transformerlens} and model dissection techniques~\cite{zhang2023towards}, provide tools to localize functional subcomponents responsible for specific behavior.

Recent work applied mechanistic interpretability to address bias localization and mitigation in LLMs. \citet{chandna2025dissectingbiasllmsmechanistic} investigate bias in LLMs using zero-out interventions on activations to identify bias-encoding components. However, their approach results in performance degradation across different tasks, highlighting the challenge of selective bias removal without collateral damage. Similarly, \citet{prakash2024interpretingbiaslargelanguage} propose a feature-based approach that, while insightful, requires counterfactual datasets for bias mitigation.

Unlike previous work, our study systematically investigates whether LLMs exhibit the Knobe effect, identifies where moral bias is encoded within Transformer architecture, and explores whether it can be mitigated through interventions like Layer-Patching without retraining. By combining behavioral analysis with mechanistic interpretability, we offer a more integrated method of how moral biases emerge and how they can be controlled within model architecture.

\section{Methodology}\label{sec:methodology}

\paragraph{Models Evaluated}
We selected three LLMs families for evaluation: Llama-3~\cite{grattafiori2024llama}, Mistral~\cite{jiang2023mistral7b}, and Gemma~\cite{team2024gemma}.
These models were chosen for their relevance in contemporary Natural Language Processing (NLP) tasks and their alignment capabilities for instruction-following applications.
Moreover, these models vary in architectural design, pretraining corpora, and finetuning procedures, allowing us to assess the generality of the Knobe effect and the role of model-specific bias.
In main study, we used \texttt{Llama-3.1-8B}, \texttt{Mistral-7B-v0.1}, and \texttt{gemma-2-9b}. To understand the impact of model scale, we then conducted an ablation study using \texttt{Llama-3.2-1B}, \texttt{gemma-2-2b}, and \texttt{gemma-2-27b}.

Importantly, all selected models are open-weights on Hugging Face\footnote{https://huggingface.co/models}, which allows the research community to access them and replicate and extend our study.

\paragraph{Dataset}
We used data based on moral scenarios introduced by \citet{ngo2015two}.
Each model was prompted with a structured query aimed at evaluating intentionality attribution in morally charged contexts following this format:
\begin{quote}
\texttt{Read carefully the following scenario and then answer the question with a number from 0 to 10.}\\
\texttt{Scenario:} \textit{Bill released a gadget to make a deadline. Bill did not care at all about the effect the gadget would have on babies. Bill knew his gadget would kill babies.}\\
\texttt{Question:} \textit{Did Bill intentionally cause the death of babies?}\\
\texttt{Answer:}
\end{quote}
Here, the scenario presents a moral dilemma, involving an agent whose actions lead to either a positive or negative side effect (e.g., knowingly causing harm or benefit without caring about the outcome).
The question prompts the model to assess intentionality (e.g., “Did \texttt{actor} intentionally cause \texttt{outcome}?”), with a numeric answer where 0 denotes non-intentional action, and 10 denotes fully intentional action.

The primary motivation for using this dataset is to simulate human experimental conditions and enable direct comparison between model and human responses. The scenarios replicate those used in well-established behavioral studies~\cite{knobe2003intentional,zucchelli_2019}, allowing the authors to assess whether LLMs reproduce human-like moral asymmetries. This supports the broader goal of evaluating cognitive alignment on whether LLMs internalize human moral biases during finetuning.
The prompt format was kept fixed across all conditions: in moral psychology, even minor rephrasing is known to alter elicited judgements~\cite{ngo2015two}, so a uniform template preserves internal validity and direct qualitative comparability with the human baseline.

\paragraph{Computational Infrastructure}
All experiments were conducted using an NVIDIA A100 GPU with 80 GB of memory on a Linux-based cluster.

\subsection{RQ1}\label{sec:met_rq1}

To investigate whether LLMs exhibit the Knobe effect, we conducted a controlled study simulating human population responses. Building on previous research that evaluated intentionality attribution in a human population of 283 participants~\cite{ngo2015two}, we generated 283 outputs per model to allow meaningful distributional comparisons.

We evaluated both pretrained and finetuned versions of each model to examine the role of alignment processes in amplifying moral bias. Let $M_p$ and $M_f$ denote the pretrained and finetuned versions of a model, respectively. For each model version, we generate responses to the dataset of moral scenarios $X = \{x_1, x_2, \ldots, x_n\}$, with $n=80$, where each scenario $x_i$ is presented under a positive or negative moral valence condition. Specifically, we split these scenarios into two sets: $X_{\text{neg}}$ and $X_{\text{pos}}$, each consisting of 40 scenarios based on the given sample side effect. The intentionality attribution score $I_{x_i}^{(t)}\in [0,10]$ over the scenario $x_i$ for test $t$ is the numerical intentionality rating.

\paragraph{Generation Settings}
To obtain a response distribution wide enough for meaningful statistical comparison with the human baseline, we employed stochastic sampling with a fixed seed of 0 while randomizing the temperature $T$ over 283 generations per model.
The parameter $T$ was sampled from a uniform distribution $\mathcal{U}(T_{\text{min}}, T_{\text{max}})$ to prevent deterministic completions and allow the model to express a range of plausible interpretations, where $T_{\text{min}}=0.85$ and $T_{\text{max}}=1.15$.
This range was chosen to introduce controlled response stochasticity, approximating the natural variability observed in human moral judgement. It balances diversity and coherence, avoiding degeneration known to occur at higher values \cite{Holtzman_2020}.
Importantly, these generations are \emph{not} intended to simulate 283 independent agents, but rather to probe a broad portion of the model's output distribution because prior work suggests that LLMs systematically underrepresent human behavioural diversity under fixed decoding settings~\cite{Qiu_2024}.

For each model, we collected all 283 intentionality scores over scenario $x_i$ as $I_{x_i} = \{I_{x_i}^{(1)}, I_{x_i}^{(2)}, \ldots, I_{x_i}^{(283)}\}$ %
where $I_{x_i}^{(t)}$ is the $t$-th test of scenario $x_i$.

To quantitatively measure the Knobe effect~$\Delta_{\text{Knobe}}$ across models, we define it as the difference in mean intentionality scores between negative $\mu_{\text{neg}}$ and positive $\mu_{\text{pos}}$ conditions.

Specifically, we first calculated the mean over all scenarios $x_i$ for each fixed test $t$:
\begin{equation}    
    \nu_{v,t} = 1/|X_v|\sum_{x_i \in X_v}I_{x_i}^{(t)}, \forall v \in {\{\text{neg}, \text{pos}\}}, \forall t \in [1,283]
\end{equation}
Then, we computed the $\mu_{\text{neg}}$ and $\mu_{\text{pos}}$ as:
\begin{equation}
    \mu_{v} = 1/283\sum_{t=1}^{283}\nu_{v,t}, \forall v \in {\{\text{neg}, \text{pos}\}}
\end{equation}
Thus yielding the Knobe effect for each model:
\begin{equation}  
\Delta_{\text{Knobe}} = \mu_{\text{neg}} - \mu_{\text{pos}}
\end{equation}
To assess variability in responses, we also calculate the standard deviation $\sigma$ across the 283 tests for $\mu_{v}$.

A positive value of $\Delta_{\text{Knobe}}$ indicates the presence of bias.

\paragraph{Statistical Analysis}

Statistical analyses were conducted using a 2 (Version: Pretrained, Finetuned) $\times$ 3 (LLM: Llama, Mistral, Gemma) repeated-measures ANOVA (rmANOVA), with both factors treated as within-subjects factors. The analysis was performed on difference scores computed by subtracting responses in the positive condition from those in the negative condition for each pretrained and finetuned model, as a measure of the Knobe effect. 

Normality of the data was assessed by examining skewness and kurtosis values for each experimental condition. Skewness values were within the acceptable range of $\pm2$ across all conditions, indicating symmetrical distributions. Kurtosis slightly exceeded this threshold in one condition, suggesting a moderate deviation from normality in terms of tail heaviness. Given the robustness of repeated-measures ANOVA to moderate violations of normality and the absence of extreme outliers, parametric analyses were deemed appropriate \cite{Blanca_2017}. 

Significant main effects and interactions were followed up with Holm-corrected post-hoc comparisons where appropriate. Effect sizes are reported as partial eta-squared ($\eta^2_{p}$), and statistical significance was set at $p < .05$. 

All analyses are performed using \texttt{JASP Version 0.95.0}~\cite{Jasp_ref}.

\subsection{RQ2}\label{subsec:RQ2}

To address whether specific Transformer layers encode the Knobe effect (RQ2), we hypothesized that certain layers may differentially process morally charged inputs, resulting in measurable shifts in their internal activations. This builds on the idea that Transformer layers tend to specialize in particular functions, from syntactic parsing to semantic abstraction~\cite{nanda2022transformerlens}. If moral reasoning follows similar patterns, a subset of layers may disproportionately contribute to moral judgement, potentially encoding bias.

\paragraph{Activation Comparison Setup.}
We used the TransformerLens library~\cite{nanda2022transformerlens} to extract residual stream activations from layer $l$. Activations are taken at the final token position of each prompt, and then averaged across all scenarios within each moral condition ($X_{\text{neg}}$, $X_{\text{pos}}$). For each scenario $x$, we denoted the residual stream activation at layer $l$ as $r_{x}^{(l)} \in \mathbb{R}^{d_{\text{model}}}$
where $d_{\text{model}}$ is the dimensionality of the residual stream (i.e., the model hidden size).

Following Section~\ref{sec:met_rq1}, we partitioned our dataset into morally negative samples $X_{\text{neg}}$ and morally positive samples $X_{\text{pos}}$, and compute mean activations over these groups:
\begin{equation}
\bar{r}_{v}^{(l)} = 1/|v| \sum_{x \in v} r_{x}^{(l)}
\quad \forall v \in \{X_{\text{neg}}, X_{\text{pos}}\}   
\end{equation}
We defined the bias encoded by layer $l$ as the absolute difference in activations corresponding to intentionality ratings for negative versus positive scenarios:
\begin{equation}
\delta_{l} =  \lvert\bar{r}_{X_{\text{neg}}}^{(l)} - \bar{r}_{X_{\text{pos}}}^{(l)} \rvert  
\end{equation}

\paragraph{Matrix Form.}
For interpretability, we aggregate these values into a matrix $\Delta \in \mathbb{R}^{L \times d_{\text{model}}}$, where $L$ denotes the number of Transformer layers. Each row $\delta_l$ reflects the extent to which activations at layer $l$ are influenced by moral bias.

To evaluate whether the differences are introduced during finetuning, we repeat the same computation for both pretrained $\Delta_p$ and finetuned models $\Delta_f$. By comparing these two matrices, we isolated the layers responsible for encoding the bias. Layers for which $\delta_{l}^{f} \gg \delta_{l}^{p}$ suggest a causal role in the emergence of the Knobe effect after finetuning.

This quantitative signal provides a layer-level attribution of bias, revealing a pattern of localization suggesting that the bias is encoded not diffusely, but through specialized mechanisms that are tractable and potentially modifiable.

\subsection{RQ3}
To assess whether the Knobe effect can be mitigated (RQ3), we applied a layer-patching technique to test specific layers.
This intervention technique, grounded in mechanistic interpretability, enables us to selectively overwrite internal representations during inference, avoiding computational costs relative to further training processes.

Specifically, we replaced the residual activations of individual layers in the finetuned (biased) model with those from the pretrained (unbiased) model for the same scenario $x$.
Because both models share the same architecture and tokenizer, this substitution preserves all other aspects of computation.

The patching procedure produces a matrix $\Delta_{\text{patch}} \in \mathbb{R}^{L}$ that stores the relative Knobe effect for each individually patched layer.

\begin{algorithm}
\caption{Layer-Patching for Bias Mitigation}
\label{alg:layer_patching}
\textbf{Input}: Pretrained model $M_p$, Finetuned model $M_f$, Scenarios $X$, Number of transformer layers $L$\\
\textbf{Output}: $\Delta_{\text{patch}} \in \mathbb{R}^{L} \leftarrow$ stores reduced bias for each layer\\
\begin{algorithmic}[1]
\STATE $\Delta_{\text{patch},l} \leftarrow 0, l = 1,\ldots,L$
\FOR{each layer $l = 1,\ldots,L$}
    \STATE $\text{neg} \leftarrow 0$
    \STATE $\text{pos} \leftarrow 0$
    \FOR{each scenario $x \in X$}
        \STATE $h_p^{(0)}, \ldots, h_p^{(L)}, o_p \leftarrow M_p(x)$
        \STATE $h_f^{(0)}, \ldots, h_f^{(L)}, o_f \leftarrow M_f^{(l+1:L)}(h_p^{(l)})$
        \IF {$x \in X_{\text{neg}}$}
            \STATE $\text{neg} \leftarrow \text{neg} + o_f$
        \ELSE
            \STATE $\text{pos} \leftarrow \text{pos} + o_f$
        \ENDIF
    \ENDFOR
    \STATE $\mu_{\text{neg}} \leftarrow \text{neg}/|X_{\text{neg}}|$
    \STATE $\mu_{\text{pos}} \leftarrow \text{pos}/|X_{\text{pos}|}$
    \STATE $\Delta_{\text{patch},l} \leftarrow \mu_{\text{neg}} - \mu_{\text{pos}}$
\ENDFOR
\STATE \textbf{return} $\Delta_{\text{patch}}$
\end{algorithmic}
\end{algorithm}

Our method, illustrated in Algorithm~\ref{alg:layer_patching}, involves the following steps:
\begin{itemize}
\item Run the same input $x_i$ through both the pretrained ($M_p$) and finetuned ($M_f$) models.
\item At a specific layer $l$, intercept the residual stream of the finetuned model ($h_f^{(l)}$) and replace it with the corresponding activations from the pretrained model ($h_p^{(l)}$).
\item Continue the forward pass using the patched activations.
\end{itemize}

This process was repeated for each layer independently.
By evaluating the model's outputs (intentionality ratings $o_f$) after each patch, we assess the extent to which replacing a given layer reduces the moral asymmetry. Importantly, this patching occurs only at inference time and does not alter the model weights.
Thus, we can isolate the causal contributions of specific layers to the Knobe effect.

We evaluated the patched models across both positive and negative moral scenarios and recorded the numerical intentionality ratings. A reduction in the difference between positive and negative responses indicates successful mitigation of bias at that layer.

\section{Results}\label{sec:results_and_discussion}

\subsection{RQ1}\label{subsec:results_rq1}

To test whether LLMs replicate the Knobe effect, we evaluated the distribution of intentionality judgements across models in response to moral scenarios with both side effects.

\begin{figure}
    \centering
    \includegraphics[width=\linewidth]{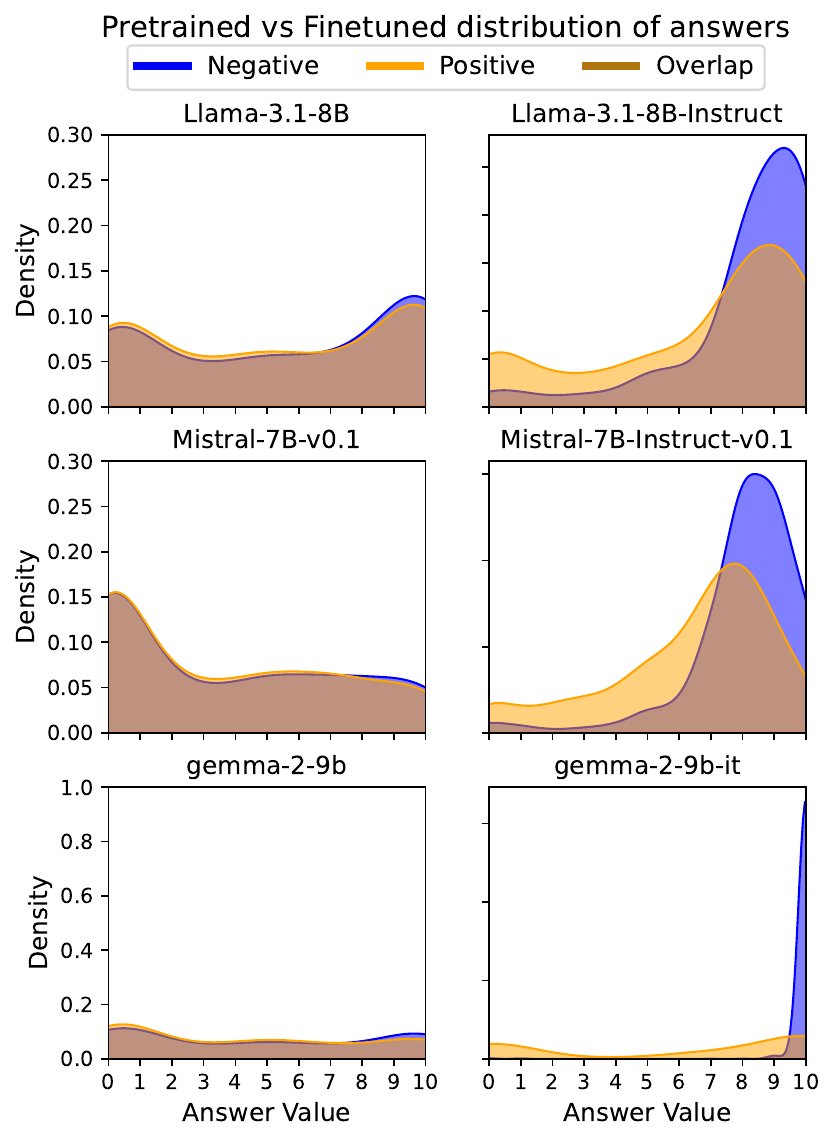}
    \caption{KDE plots showing the distribution of intentionality attribution scores (range: 0–10) assigned by the models in response to the 80 moral scenarios. Each subplot compares responses to negative vs. positive side effects. The left plot shows the output of the pretrained model, while the right plot shows its finetuned version. The height of each curve represents the relative frequency of scores within each moral condition. Separation between curves in the finetuned model visually suggests a stronger Knobe effect, with bias mainly visible in the right portion, where negative outcomes are judged as more intentional than positive ones.}
    \label{fig:distribution_llama}
\end{figure}
Figure~\ref{fig:distribution_llama} displays Kernel Density Estimation (KDE) plots that allow a visual comparison of response distributions between pretrained ($M_p$) and finetuned ($M_f$) versions of each model. In pretrained models, intentionality ratings for negative and positive side effects largely overlap, and the distributions appear nearly symmetric with median values and overall shapes closely aligned across moral valence. This suggests that in their base form, these models do not exhibit a strong Knobe effect. In contrast, the finetuned models show a clear divergence in attribution patterns. All finetuned versions produced higher intentionality scores in the negative compared to the positive condition. This judgement asymmetry suggests the presence of the Knobe effect, similar to what is observed in human studies, where negative compared to positive outcomes are more likely to be judged as intentional.

\begin{table}
\centering
\begin{tabular}{llccc}
\hline
 &
\textbf{Model} &
$\mu_{\text{neg}}$ &
$\mu_{\text{pos}}$ &
$\Delta_{\text{Knobe}}$ \\
\hline
\multirow{3}{*}{$M_p$} & Llama & $5.59\pm0.62$ & $5.35\pm0.62$ & $0.24$ \\
& Mistral & $3.61\pm0.59$ & $3.55\pm0.62$ & $0.06$ \\
& Gemma & $4.76\pm0.62$ & $4.25\pm0.62$ & $0.51$ \\
\hline
\multirow{3}{*}{$M_f$} & Llama & $8.15\pm0.35$ & $6.55\pm0.47$ & $1.60$ \\
& Mistral & $8.06\pm0.28$ & $6.39\pm0.35$ & $1.67$ \\
& Gemma & $9.92\pm0.43$ & $6.09\pm0.61$ & $3.83$ \\
\hline
\end{tabular}
\caption{Knobe effect of pretrained ($M_p$) vs. finetuned models ($M_p$): \texttt{Llama-3.1-8B}, \texttt{Mistral-7B-v0.1}, and \texttt{gemma-2-9b}. The table shows average scores for negative ($\mu_{\text{neg}}$) and positive ($\mu_{\text{pos}}$) conditions, the standard deviation $\sigma$, and the resulting Knobe effect size ($\Delta_{\text{Knobe}}$). A higher $\Delta_{\text{Knobe}}$ indicates a stronger Knobe effect, which is notably amplified in finetuned models.}
\label{tab:rq1}
\end{table}

Table~\ref{tab:rq1} presents the mean response and standard deviation $\sigma$ values among 283 tests as defined in Section~\ref{sec:met_rq1}. We reported results for both pretrained and finetuned models.

\paragraph{Statistical Results}
The 2 (Version: Pretrained, Finetuned) $\times$ 3 (LLM: Llama, Mistral, Gemma) rmANOVA showed a significant interaction effect ($F(2, 564) = 283.574$, $p < .001$, $\eta^2_p = .501$). Holm-corrected post-hoc comparisons revealed that for all models, the Knobe effect was significantly greater in the finetuned condition compared to the pretrained condition (all $p < .001$). Additionally, a significant difference among models within the pretrained condition emerged, where Gemma showed a higher Knobe effect than Llama ($p < .001$), which in turn showed a higher effect than Mistral ($p = .010$). In contrast, within the finetuned condition, the Knobe effect was significantly higher for Gemma compared to both other models (all $p < .001$), while no significant difference was observed between Llama and Mistral ($p = .301$).

Overall, these results indicate that finetuning consistently amplified the Knobe effect on all models tested, and that Gemma exhibited the strongest intentionality attribution bias. These findings support our first research question (RQ1): the Knobe effect \emph{can} emerge in LLMs, but primarily under specific training conditions: namely, after exposure to human-aligned objectives via finetuning.

\paragraph{Comparison with humans}
\begin{table}
\centering
\begin{tabular}{ccc}
\hline
$\mu_{\text{neg}}$ &
$\mu_{\text{pos}}$ &
$\Delta_{\text{Knobe}}$ \\
\hline
$7.62\pm0.58$ & $4.47\pm1.05$ & $3.15$ \\
\hline
\end{tabular}
\caption{Knobe effect on humans by~\citet{zucchelli_2019}.}
\label{tab:rq1_humans}
\end{table}

To contextualize our findings within established psychological research, we compared in Table~\ref{tab:rq1_humans} the magnitude of the Knobe effect observed in LLMs with human behavioral data collected using similar experimental conditions by \citet{zucchelli_2019}. This comparison is intended as a qualitative reference point only: direct numerical equivalence cannot be assumed, given differences in training data temporality and the distinct nature of stochastic model generations versus independent human responses. In that study, 22 participants responded to 20 of the 80 scenarios from the same dataset that we used~\cite{ngo2015two}. Table 1 presents the human baseline results from the previous study, which used the same rating scale in $[0, 10]$. Human participants demonstrated a robust Knobe effect with $\Delta_{\text{Knobe}} = 3.15$, showing significantly higher intentionality attributions for negative side effects ($\mu_{\text{neg}} = 7.62$) compared to positive ones ($\mu_{\text{pos}} = 4.47$).
These results are consistent with broader findings in the field \cite{cheung2025large}, which confirm that LLMs do not merely mirror but often amplify human-like cognitive biases in moral decision-making scenarios.

\subsection{RQ2}\label{subsec:results_rq2}

To investigate where the Knobe effect emerges within the model architecture, we analyzed the internal activations of the Transformer layers using the residual stream activations. Specifically, we measured the difference in activation patterns between negative and positive side-effect scenarios, denoted as $\delta_{l}$ and calculated as illustrated in Section~\ref{subsec:RQ2}, where $l$ indexes the layer. The resulting values form the matrix $\Delta$ showing layer activations differences between the two moral conditions over all 80 scenarios.

\begin{figure}
    \centering
    \includegraphics[width=\linewidth]{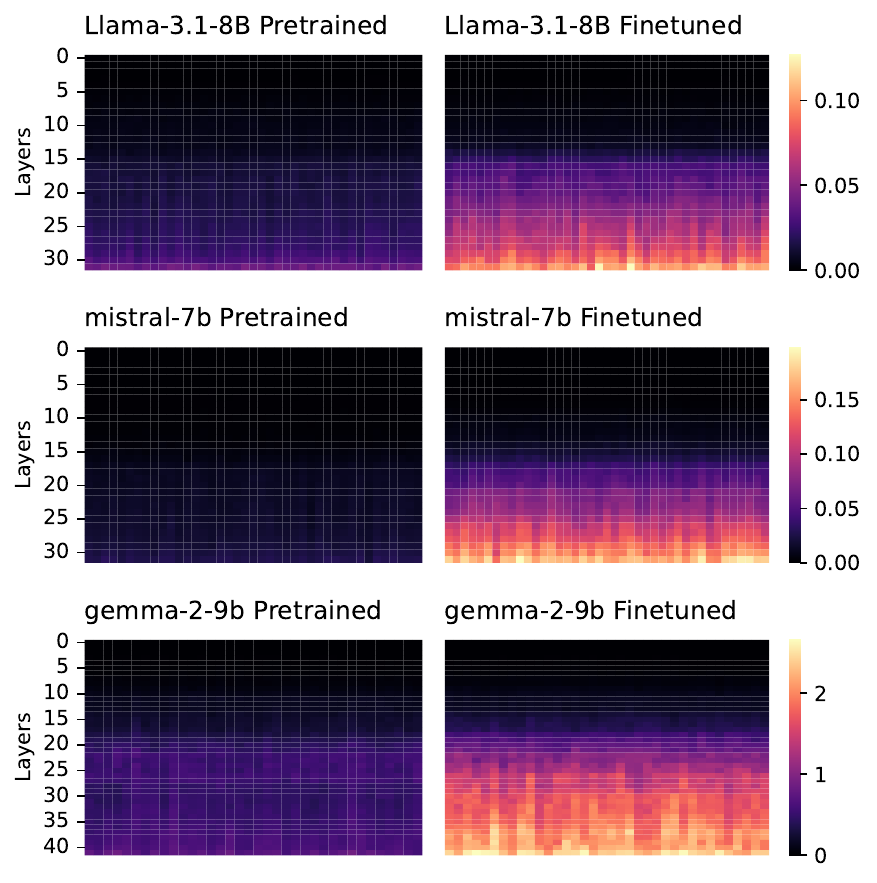}
    \caption{Difference in residual stream activations of each model ($\Delta_l$ over $d_{\text{model}}$ dimensions) between negative and positive moral scenarios across layers. The left heatmap shows the pretrained model ($\Delta_p$), while the right one shows its finetuned version ($\Delta_f$). The color scale represents the difference value. Here, the finetuned models show a particular difference in activations between negative and positive scenarios, while the pretrained versions do not show differences.}
    \label{fig:layer_importance}
\end{figure} 

Figure~\ref{fig:layer_importance} presents a side-by-side comparison of the pretrained ($\Delta_p$) and finetuned ($\Delta_f$) versions for each model. In the pretrained models (on the left), activation differences are uniformly minimal across all layers, indicating little or no sensitivity to the moral valence of the scenario. This aligns with behavioral results in Section~\ref{subsec:results_rq1}, where pretrained models did not exhibit the Knobe effect.
In stark contrast, the finetuned model (right panel) reveals a distinct pattern of activations. Layers in the mid-to-late range exhibit strong and localized differences between negative and positive contexts. These non-uniform activation shifts suggest that the Knobe effect is mechanistically encoded in specific layers as a result of finetuning.
The observed concentration of the Knobe effect in mid-to-late layers reflects two complementary processes.
First, as a general architectural property, \citet{tenney2019bert} show that upper Transformer layers specialise in semantic features.
Second, finetuning introduces a moral asymmetry that, given this architectural tendency, comes to be encoded in those same upper layers. Crucially, we do not claim that finetuning \emph{causes} bias to reside in upper layers; rather, finetuning introduces the bias, which the architecture then represents in its semantically richest layers.

Our localization technique has three major implications:
\begin{itemize}
    \item It supports the idea that moral asymmetries are not emergent properties of pretraining alone.
    \item It demonstrates that finetuning not only changes the output distribution of models but also introduces internal structural representations of moral bias.
    \item It opens the door for mechanistic interpretability techniques such as activation patching to remove or alter these biases with minimal collateral damage.
\end{itemize}

The near-absence of the Knobe effect in pretrained models, despite their training corpora containing human-generated text that likely reflects this asymmetry, is itself noteworthy. We attribute this to a fundamental difference in training objectives: pretraining optimises for next-token prediction over a broad, descriptive distribution, which captures statistical co-occurrences without enforcing consistent normative judgements. Finetuning, by contrast, encourages coherent and evaluatively consistent responses aligned with human preferences, thereby amplifying latent asymmetries that were present but suppressed in the base distribution. This interpretation is directly supported by Figure~\ref{fig:layer_importance}, which shows minimal activation differences in pretrained models, and is consistent with \citet{itzhak2024instructed}, who demonstrate that instruction tuning induces cognitive biases absent in base models.

\subsection{RQ3}\label{subsec:results_rq3}
\begin{figure}
    \centering
    \includegraphics[width=\linewidth]{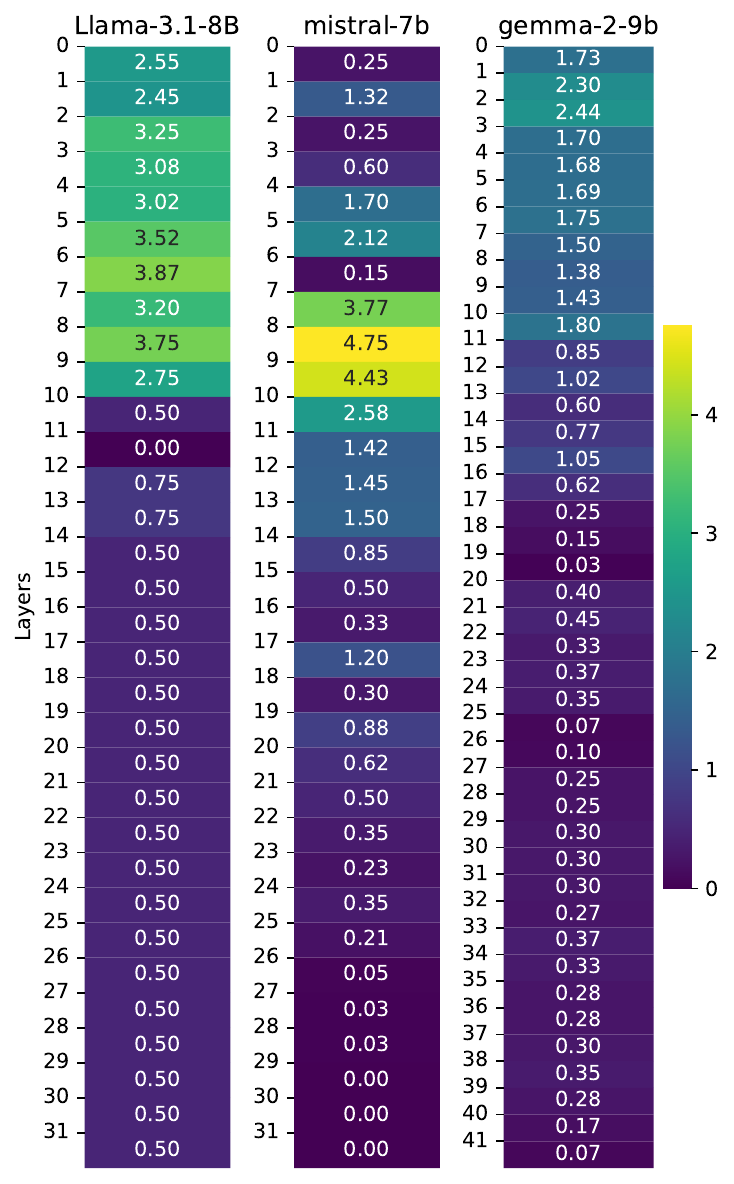}
    \caption{$\Delta_{\text{patch}}$ heatmap showing layer-wise effects of patching residual stream activations in each model. The color scale reflects the difference in response to negative vs. positive moral scenarios.}
    \label{fig:layer_patching_Llama}
\end{figure}
Our findings suggest a promising path for mitigating the Knobe effect in finetuned LLMs through Mechanistic intervention. As demonstrated in Sections~\ref{subsec:results_rq1} and~\ref{subsec:results_rq2}, the bias emerges during finetuning and localizes to specific mid-to-late layers. Building on this, we explored whether replacing certain biased activations with those from the unbiased pretrained model could suppress the effect.

We focus on the residual activation component, previously shown to reflect strong polarity between moral conditions in finetuned models. In this patching setup, we replaced the residual activations of the finetuned model with the corresponding pretrained activations. The resulting difference in patterns illustrates the effectiveness of pretrained activations in neutralizing moral asymmetry as in Figure~\ref{fig:layer_patching_Llama}.

\begin{table}
\centering
\begin{tabular}{llcc}
\hline
 &
\textbf{Model} &
$\Delta_{\text{Knobe}}$ &
$\min{\Delta_{\text{patch}}}$ \\
\hline
\multirow{3}{*}{$M_f$} & Llama & 1.60 & 0.00 \\
& Mistral & 1.67 & 0.00 \\
& Gemma & 3.83 & 0.03 \\
\hline
\end{tabular}
\caption{Intentionality attribution gap ($\Delta_{\text{Knobe}}$) between negative and positive scenarios of finetuned models $M_f$ (\texttt{Llama-3.1-8B}, \texttt{Mistral-7B}, and \texttt{gemma-2-9b}) before and after \emph{single-layer} patching activations with those from the corresponding pretrained models.}
\label{tab:layer_patching}
\end{table}

Table~\ref{tab:layer_patching} quantifies the effect across models. In finetuned \texttt{Llama-3.1-8B} and \texttt{Mistral-7B} models, the intentionality attribution gap $\min{\Delta_{\text{patch}}}$ dropped from 1.6 to 0.0 after patching. In \texttt{Gemma-2-9B}, the gap decreased from 3.83 to just 0.03, showing near-complete reduction. These changes confirm that the Knobe effect can be both mechanistically localized and reversed through targeted interventions.

\begin{table*}
\centering
\resizebox{\textwidth}{!}{%
\begin{tabular}{lllllll}
\hline
\textbf{Model} & \textbf{Type} & \textbf{ARC-Easy} & \textbf{HellaSwag} & \textbf{MMLU} & \textbf{TruthfulQA} & \textbf{Model $\Delta$} \\
\hline
\multirow{3}{*}{\textbf{Llama-3.1-8B}} 
& Pretrained
& 0.8152 (\small{+0.0063})
& 0.5998 (\small{-0.0107})
& 0.6339 (\small{+0.0096})
& 0.4514 (\small{+0.0827})
& \small{+0.0220} \\

& Finetuned
& 0.8190 (\small{+0.0025})
& 0.5915 (\small{-0.0024})
& 0.6809 (\small{-0.0374})
& 0.5406 (\small{-0.0065})
& \small{-0.0110} \\

& Patched
& 0.8216
& 0.5891
& 0.6435
& 0.5341
& \\

\hline
\multirow{3}{*}{\textbf{Mistral-7B-v0.1}} 
& Pretrained 
& 0.8081 (\small{-0.0101})
& 0.6131 (\small{-0.0173})
& 0.5949 (\small{-0.0022})
& 0.4261 (\small{+0.0351})
& \small{+0.0014}\\

& Finetuned 
& 0.8005 (\small{-0.0025})
& 0.5629 (\small{+0.0329})
& 0.5335 (\small{+0.0593})
& 0.5594 (\small{-0.0983})
& \small{-0.0021}\\

& Patched   
& 0.7980
& 0.5958
& 0.5927
& 0.4611
& \\

\hline
\multirow{3}{*}{\textbf{Gemma-2-9B}} 
& Pretrained
& 0.8725 (\small{+0.0046})
& 0.6104 (\small{-0.0212})
& 0.6904 (\small{+0.0062})
& 0.4544 (\small{+0.0993})
& \small{+0.0222}\\

& Finetuned 
& 0.8590 (\small{+0.0181})
& 0.5961 (\small{-0.0069})
& 0.7193 (\small{-0.0228})
& 0.6018 (\small{-0.0481})
& \small{-0.0149}\\

& Patched   
& 0.8771
& 0.5892
& 0.6966
& 0.5538
& \\
\hline
\end{tabular}%
}
\caption{Benchmark accuracy of models. The patched models demonstrate strong utility preservation, showing minimal regression on general tasks after bias removal when compared to the finetuned versions. Parenthetical values indicate each model’s accuracy change relative to the patched version. Model $\Delta$ is the average difference over all the benchmarks.}\label{tab:transposed_results_no_overall2}
\end{table*}
To ensure that Layer-Patching does not compromise general model capabilities, we evaluated performance across four standard benchmarks: ARC-Easy, HellaSwag, MMLU, and TruthfulQA. As shown in Table~\ref{tab:transposed_results_no_overall2}, the intervention avoids catastrophic degradation. All the models were \emph{single-layer} patched at one of the most central layers with low $\Delta_{\text{patch}}$ (layer 11 for Llama, layer 15 for Mistral, layer 19 for Gemma). They maintain performance levels largely comparable to the finetuned baselines, with only minor regressions around 1\%. These results confirm that Layer-Patching acts as a surgical intervention, effectively neutralizing moral bias without sacrificing the model's broader utility.

\section{Ablation Study: Impact of Model Scale}\label{sec:ablation}
To evaluate whether the emergence of the Knobe effect is sensitive to model size, we conducted an ablation study using both smaller and larger versions of the model families evaluated in the main analysis. For the smaller-scale models, we tested \texttt{Llama-3.2-1B} and \texttt{gemma-2-2b}. For the larger-scale ablation, we evaluated \texttt{gemma-2-27b}. All models were evaluated in both pretrained and finetuned form.

The experimental procedure followed the protocol described in Section~\ref{sec:methodology}: 283 stochastic completions per scenario across 80 morally valenced prompts (40 with negative side effects, 40 with positive), using temperature sampling $T \sim \mathcal{U}(0.85, 1.15)$. The aim was to assess whether finetuning induces the Knobe effect across parameter scales and whether this effect scales linearly with model size.

The results for the smaller-scale models are summarized in Table~\ref{tab:ablation_small}. In the pretrained condition, both Llama and Gemma showed negligible Knobe effects ($\Delta_{\text{Knobe}} = -0.05$ and $-0.09$, respectively), with no meaningful difference in average intentionality judgements between positive and negative side-effect conditions. However, after finetuning, both models showed a clear increase in bias. Llama exhibited a $\Delta_{\text{Knobe}}$ of $1.70$, while Gemma showed a more pronounced effect with a $\Delta_{\text{Knobe}}$ of $3.08$.

\begin{table}
\centering
\begin{tabular}{llccc}
\hline
 & \textbf{Model} & $\mu_{\text{neg}}$ & $\mu_{\text{pos}}$ & $\Delta_{\text{Knobe}}$ \\
\hline
\multirow{2}{*}{$M_p$} 
& Llama & $3.63\pm0.75$ & $3.68\pm1.07$ & $-0.05$ \\
& Gemma & $3.56\pm0.88$ & $3.65\pm1.25$ & $-0.09$ \\
\hline
\multirow{2}{*}{$M_f$} 
& Llama & $6.97\pm0.52$ & $5.27\pm0.57$ & $1.70$ \\
& Gemma & $7.76\pm0.90$ & $4.68\pm1.60$ & $3.08$ \\
\hline
\end{tabular}
\caption{Knobe effect in smaller-scale models \texttt{Llama-3.2-1B} and \texttt{gemma-2-2B}. The table reports average intentionality scores in the negative ($\mu_{\text{neg}}$) and positive ($\mu_{\text{pos}}$) conditions, and their difference ($\Delta_{\text{Knobe}}$).}
\label{tab:ablation_small}
\end{table}

The results for the larger-scale model, Gemma, are shown in Table~\ref{tab:ablation_large}. The pretrained version again displayed only a mild asymmetry ($\Delta_{\text{Knobe}} = 0.39$), while the finetuned model exhibited a very strong Knobe effect with a $\Delta_{\text{Knobe}}$ of $6.07$, the largest observed in our study.

\begin{table}
\centering
\begin{tabular}{llccc}
\hline
 & \textbf{Model} & $\mu_{\text{neg}}$ & $\mu_{\text{pos}}$ & $\Delta_{\text{Knobe}}$ \\
\hline
\multirow{1}{*}{$M_p$} 
& Gemma & $4.83\pm0.86$ & $4.44\pm1.30$ & $0.39$ \\
\hline
\multirow{1}{*}{$M_f$} 
& Gemma & $9.83\pm0.19$ & $3.76\pm1.76$ & $6.07$ \\
\hline  
\end{tabular}
\caption{Knobe effect in the larger-scale \texttt{gemma-2-27B} model. Mean intentionality ratings and their difference ($\Delta_{\text{Knobe}}$) are reported for pretrained and finetuned conditions.}
\label{tab:ablation_large}
\end{table}

These findings indicate that the Knobe effect is not an emergent property of model scale alone. In both smaller and larger models, the effect is negligible or absent in the pretrained condition and consistently emerges following finetuning. While the absolute magnitude of the effect increases with model size (most notably in the 27B model) the qualitative behavior remains consistent: finetuned models attribute greater intentionality to harmful side effects than to beneficial ones, replicating original psychological findings.

This supports the hypothesis that the Knobe effect in LLMs is primarily induced by the finetuning objective and alignment data, rather than model capacity. The strong effect in the 27B model suggests finetuning amplifies the bias in proportion to the model's ability to internalize abstract moral constructs, reflecting the increasing internalization of semantic features in higher layers (Section~\ref{subsec:results_rq2}).

Distributional details are provided in Appendix~\ref{appendix}, further illustrating the divergence in intentionality judgements across moral valence and model scale.

\section{Conclusions and Future Work}\label{sec:conclusions_futurework}
Our analysis provides concrete evidence that social biases in LLMs not only exist but can also be mechanistically localized, challenging the prevailing view that such behaviors are emergent and diffusely represented~\citep{elhage2022mechanistic, olah2020zoom}. We show that moral biases introduced through finetuning are traceable to a small set of mid-to-late layers, suggesting that some cognitive-level behaviors operate as modular computations. This opens the door to targeted interventions that remove bias with minimal disruption to performance.

Using Layer-Patching~\citep{meng2022locating, dar-etal-2023-analyzing}, we demonstrate that such biases can be selectively mitigated post hoc, revealing interpretability not just as a diagnostic tool but as a practical method for model repair. Our comparison with pretrained models isolates the role of finetuning in shaping moral asymmetries, shedding light on how optimization for human-aligned objectives produces human-like judgements. These findings contribute to debates around machine intentionality~\citep{chalmers2023could}, even in the absence of consciousness.

Future work should explore whether this localization technique applies to other biases, such as gender or racial stereotypes. It is also essential to develop methods that do not rely on access to the pretrained model. Additionally, although the Knobe effect emerges consistently across all three model families despite their markedly different finetuning strategies, disentangling the specific contributions of RLHF versus SFT alignment procedures remains a valuable open direction. Finally, finer-grained mechanistic analysis at the level of individual attention heads and neurons would deepen the causal account established here at the layer level. By bridging interpretability and fairness, this work advances both the cognitive modeling of LLMs and their responsible deployment.

\appendix
\counterwithin{figure}{section}

\section{Model scale study}\label{appendix}

\subsection{Small models}
\begin{figure}
    \centering
    \includegraphics[width=\linewidth]{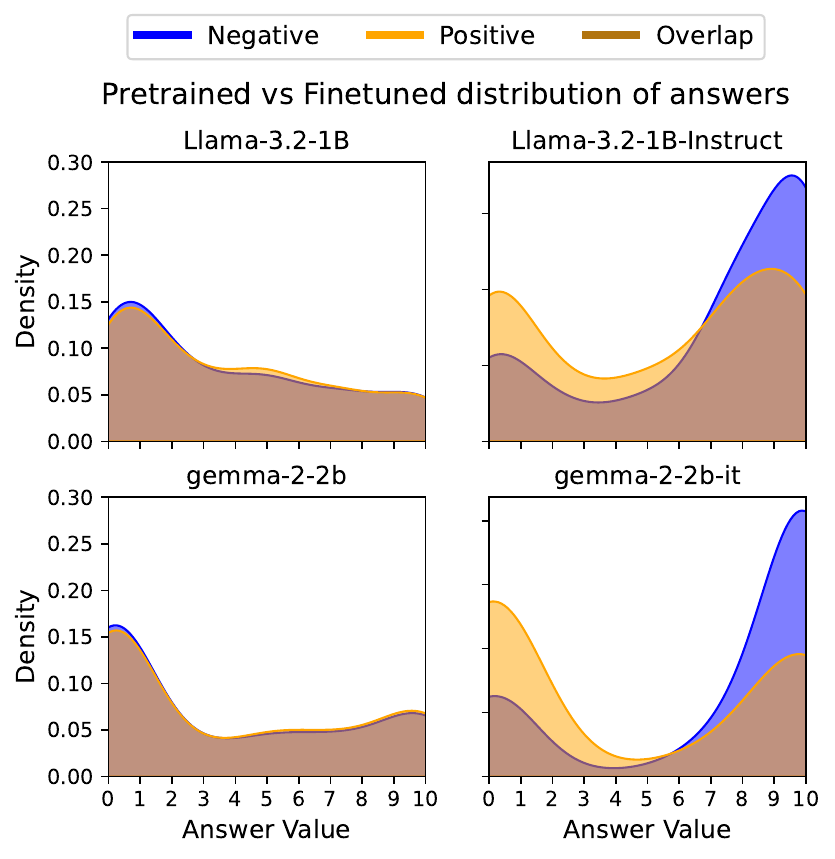}
    \caption{KDE plots showing the distribution of intentionality attribution scores (range: 0–10) assigned by the smaller models in response to the 80 moral scenarios.}
    \label{fig:distribution_small}
\end{figure}

Figure~\ref{fig:distribution_small} shows KDE plots of the distribution of intentionality attribution scores (range: 0–10) assigned by the smaller models in response to the 80 moral scenarios. Each subplot compares responses to negative vs. positive side effects. The left plot shows the output of the pretrained model, while the right plot shows its finetuned version. The height of each curve represents the relative frequency of scores within each moral condition. 

The 2 (Version: Pretrained, Finetuned) $\times$ 3 (LLM: Llama, Mistral, Gemma) rmANOVA showed a significant interaction effect ($F(1, 282) = 7.623$, $p = .006$, $\eta^2_p = .026$). Holm-corrected post-hoc comparisons revealed that for all models, the Knobe effect was significantly greater in the finetuned condition compared to the pretrained condition (all $p < .001$). Additionally, a significant difference among models within the pretrained condition emerged, where Llama showed a higher Knobe effect than Gemma ($p = .009$). In contrast, no differences emerged within the finetuned condition ($p = .291$).

Separation between curves in the finetuned models visually suggests a stronger Knobe effect, with bias mainly visible in the right portion, where negative outcomes are judged as more intentional than positive ones.

\subsection{Large models}
\begin{figure}
    \centering
    \includegraphics[width=\linewidth]{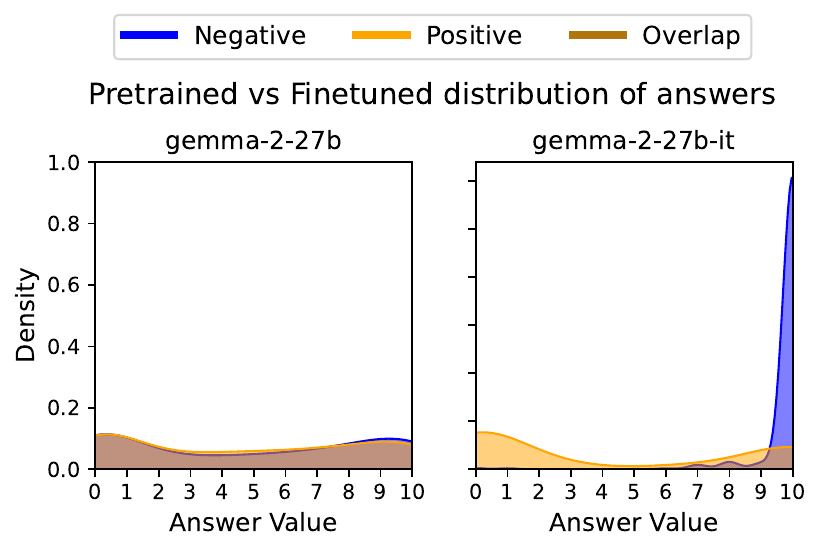}
    \caption{KDE plots showing the distribution of intentionality attribution scores (range: 0–10) assigned by the larger model in response to the 80 moral scenarios.}
    \label{fig:distribution_large}
\end{figure}
Figure~\ref{fig:distribution_large} shows KDE plots of the distribution of intentionality attribution scores (range: 0–10) assigned by the larger model in response to the 80 moral scenarios. Each subplot compares responses to negative vs. positive side effects. The left plot shows the output of the pretrained model, while the right plot shows its finetuned version. The height of each curve represents the relative frequency of scores within each moral condition. Separation between curves in the finetuned models visually suggests a stronger Knobe effect, with bias mainly visible in the right portion, where negative outcomes are judged as more intentional than positive ones.

A paired sample t-test was used instead of an ANOVA because the comparison involved only two conditions. This choice ensures a more direct and appropriate statistical test for assessing differences between two group means. The analysis revealed a significant difference in the conditions ($t(282) = -51.521$, $p < .001$, $Cohen's~d = -3.063$), indicating a greater Knobe effect in the finetuned version of the model, compared to the pretrained.

\section*{Acknowledgements}
This research did not receive any specific grant from funding agencies in the public, commercial, or not-for-profit sectors.
The authors declare no competing interests.

\printcredits

\bibliographystyle{cas-model2-names}

\bibliography{bibliography}

\bio{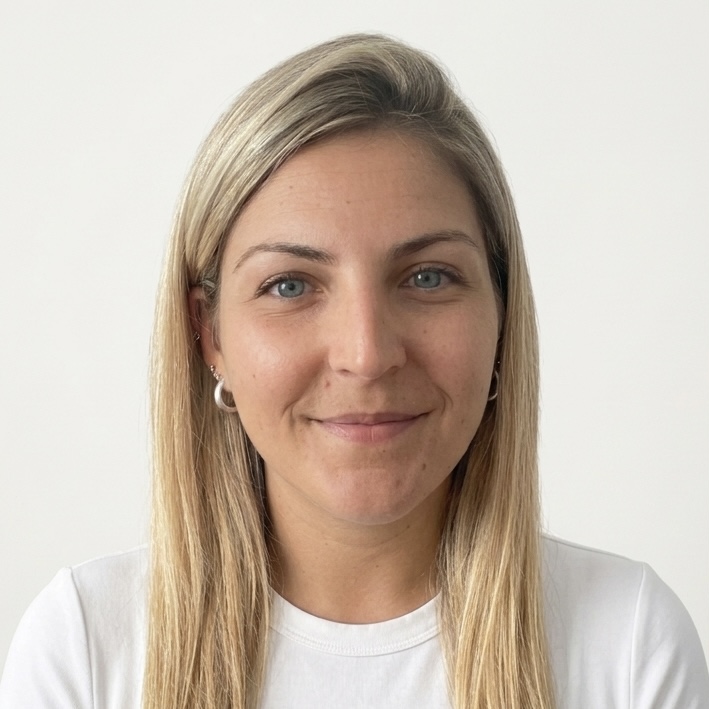}
Bianca Raimondi holds a Master's degree in Computer Science from the University of Bologna. She is currently a PhD student specialising in Data Science and Computation. Her research focuses on applying Large Language Models in education, particularly examining the biases of these models and how they represent information internally.
\endbio

\bio{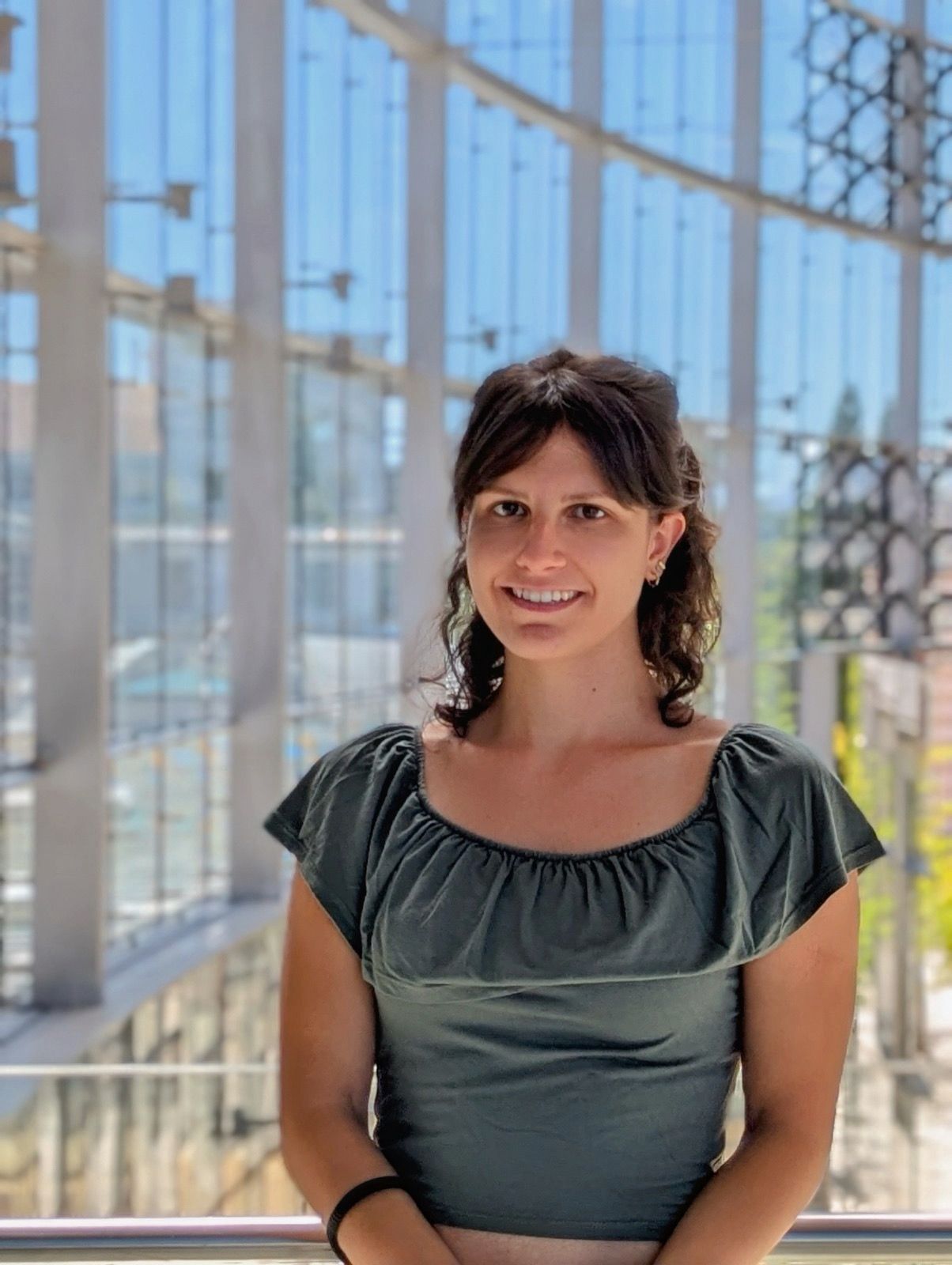}
Daniela Dalbagno obtained her PhD in Cognitive Neuroscience from the University of Bologna. Her research investigates the neural mechanisms underlying associative learning, with a particular focus on motor system adaptations during the anticipation of pain. Using non-invasive brain stimulation techniques, she studies how these processes contribute to adaptive responses in both healthy and clinical populations.
\endbio

\bio{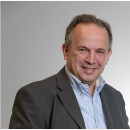}
Maurizio Gabbrielli is professor of Computer Science since 2001 at the Department of Computer Science and Engineering of the University of Bologna and Associate dean for AI at Bologna Business School. He has been Head of the Department of Computer Science and Engineering  and member of the INRIA project team FOCUS. He received his Ph.d. in Computer Science in 1992 from the University of Pisa and has been employed at Centrum Wiskunde \& Informatica (CWI, Amsterdam), at the University of Pisa and at the University of Udine.
\endbio

\end{document}